\documentclass[runningheads]{llncs}

\usepackage{tikz}
\usepackage{amsmath}
\usepackage{subcaption}
\usepackage{booktabs}
\usepackage{siunitx}
\usepackage{graphicx}
\usepackage{multirow}
\usepackage{float}
\newcommand{\mpara}[1]{\medskip\noindent{\bf #1}}
\usepackage{url}
\usepackage{xspace}
\usepackage{textcomp}
\usepackage{amsfonts}
\usepackage{filecontents}
\usepackage{algorithm2e}
\RestyleAlgo{ruled}

\counterwithin*{RQ}{section} %reset for each section

\usepackage[show]{notes-alt}

\begin{document}
%-------------------------------------------------------------------------------

\newcommand{\numc}{$\mathbf{c}$}
\newcommand{\Xst}{$\mathbf{X^*}$}
\newcommand{\ei}{$\mathcal{E}$}
\newcommand{\ALap}{$\mathbf{L}$}
\newcommand{\Y}{$\mathbf{Y}$}
\newcommand{\Xr}{$\mathbf{X^*_{R}}$}
\newcommand{\nsamp}{$\mathbf{n}$}
\newcommand{\Xsamp}{$\mathbf{X}$}
\newcommand{\Xinf}{$\mathbf{X^*_{inferred}}$}
\newcommand{\mmissing}{$\mathbf{m}$}
\newcommand{\CS}{$\mathbf{CS}$}
\newcommand{\CSsmall}{$\mathbf{cs}$}
\newcommand{\CSth}{$\mathbf{cs_{threshold}}$}
\newcommand{\CSmean}{$\mathbf{cs_{mean}}$}
\newcommand{\bsamp}{$\mathbf{b}$}
\newcommand{\SAA}{$\mathbf{SAA}$}
\newcommand{\RAA}{$\mathbf{RAA}$}
\newcommand{\phiInd}{$\mathbf{\phi}$}
\newcommand{\KKNN}{$\mathbf{K}$}
\newcommand{\KxKprop}{$K_{x}$}
\newcommand{\KyKprop}{$K_{y}$}
\newcommand{\budgetX}{$\epsilon_{X}$}
\newcommand{\budgetY}{$\epsilon_{Y}$}
\newcommand{\iter}{$\mathbf{iter}$}

\newcommand{\fpmaattack}{\textsc{Fp-ma}\xspace}
\newcommand{\rimaattack}{\textsc{Ri-ma}\xspace}
\newcommand{\riattack}{\textsc{Ri}\xspace}
\newcommand{\fpattack}{\textsc{Fp}\xspace}
\newcommand{\saattack}{\textsc{Sa}\xspace}
\newcommand{\settingone}{\textsc{Setting-1}\xspace}
\newcommand{\settingtwo}{\textsc{Setting-2}\xspace}

% make title bold and 14 pt font (Latex default is non-bold, 16 pt)
\title{Does Black-box Attribute Inference Attacks on Graph Neural Networks Constitute Privacy Risk?}
% \thanks{This work is, in part, funded by the Lower Saxony Ministry of Science and Culture under grant no. ZN3491 within the Lower Saxony ``Vorab'' of the Volkswagen
% Foundation and supported by the Center for Digital
% Innovations (ZDIN), and the Federal Ministry of Education and Research (BMBF), Germany, under the project LeibnizKILabor (grant no. 01DD20003).}

\titlerunning{Does Black-box AIA on GNNs Constitute a Privacy Risk?}

\author{Iyiola E. Olatunji\inst{1} \and
Anmar Hizber\inst{2} \and
Oliver Sihlovec\inst{3} \and
Megha Khosla\inst{3}}
\authorrunning{Olatunji et al.}
\institute{L3S Research Center, Leibniz University, Hannover, Germany \\ \email{iyiola@l3s.de} \and
Leibniz University, Hannover, Germany
\and
TU Delft, Netherlands}

\maketitle

\begin{abstract}
Graph neural networks (GNNs) have shown promising results on real-life datasets and applications, including healthcare, finance, and education. However, recent studies have shown that GNNs are highly vulnerable to attacks such as membership inference attack and link reconstruction attack. Surprisingly, attribute inference attacks has received little attention. 
In this paper, we initiate the first investigation into attribute inference attack where an attacker aims to infer the sensitive user attributes based on her public or non-sensitive attributes. We ask the question whether black-box attribute inference attack constitutes a significant privacy risk for graph-structured data and their corresponding GNN model. We take a systematic approach to launch the attacks by varying the adversarial knowledge and assumptions.
Our findings reveal that when an attacker has black-box access to the target model, GNNs generally do not reveal significantly more information compared to missing value estimation techniques. Code is available. \\

\keywords{Attribute Inference  Attack \and Privacy Risk \and Graph Neural Network.}
\end{abstract}

% =====================
\section{Introduction}

Several real-world data can be modeled as graphs. For instance, graphs are widely used to represent interactions between biological entities  \cite{PKFLGGR18,dong2022mucomid,olatunji2022review}, model the interaction between users in social networks \cite{olatunji2022private,kipf2017semi} and for designing recommendation systems~\cite{fan2019graph,wu2020graph}. 
Graph neural networks (GNNs) are a type of machine learning model that are specifically designed to handle graph-structured data. They have demonstrated effectiveness in diverse graph-based learning tasks, including node classification, link prediction, and community detection. GNNs leverage recursive aggregation of node information from neighboring nodes to generate informative graph representations \cite{kipf2017semi}. However, despite their usefulness, GNNs can pose privacy threats to the data they are trained on. Multiple studies have shown that GNNs are more vulnerable to privacy attacks than traditional machine learning methods.
These attacks include membership inference attacks \cite{emmanuelMIA,duddu2020quantifying}, link stealing attacks \cite{he2021stealing,zhang2021graph,olatunji2022private}, backdoor attacks \cite{zhang2020backdoor}, and adversarial attacks \cite{wu2019adversarial,zugner2018adversarial}.
One main reason for the high vulnerability of GNNs to attacks is their use of graph topology during training, which can lead to the leakage of sensitive information \cite{emmanuelMIA,olatunji2021releasing}. However, attribute inference attacks (AIA) have been under-explored for GNNs. In AIA, the attacker's goal is to infer the sensitive attribute value of a node via access to the target model. This poses a severe privacy risk. For instance, if a health insurance company knows the disease status of a potential client, they may discriminate against them and increase their insurance premium. We take the first step of systematically investigating the privacy risks posed by AIA on GNNs  under the practical black-box access assumptions.

As machine learning as a service (MLaaS) becomes more prevalent and GNNs are used in privacy-sensitive domains, it is essential to consider the privacy implications of black-box attribute inference attacks on GNNs. In this scenario, a user sends data to the trained model via an API, and receives a predictions. The user does not have access to the internal workings of the model. Motivated by this, we ask the question: \textit{what is the privacy implication of black-box attribute inference attack on GNNs?}. To investigate this issue, we construct several attacks in a practical scenario where an attacker has black-box access to the trained model.

We develop two attribute inference attack (AIA) methods, namely the \textit{attribute inference attack via repeated query of the target model} (\fpmaattack) and the \textit{feature propagation-only attribute inference attack} (\fpattack). In the \fpattack attack, we reconstruct the missing sensitive attributes by updating the attribute with the attribute value of the neighboring node via a feature propagation algorithm\cite{rossi2021unreasonable}.
On the other hand, the \fpmaattack attack employs a \textit{feature propagation} algorithm iteratively for each candidate node (nodes with sensitive attributes). It queries the target model with the estimated attribute and outputs a model confidence, which is then compared to a threshold to determine whether the inferred attribute is the true attribute. Additionally, we propose a \textit{shadow-based attribute inference attack} (\saattack) that assumes an attacker has access to a shadow dataset and a shadow model, similar to the target model.

The contributions of this paper can be summarized as follows: (i) we develop two black-box attribute inference attack on GNNs and a relaxed shadow attack. (ii) while most AIA focus on inferring single binary attributes, our attacks go beyond these limitations. Our approach enables the inference of both single or multiple binary attributes, as well as continuous attribute values.
(iii) through experimentation and extensive discussion, we show that having black-box access to a trained model may not result in a greater privacy risk than using missing value estimation techniques in a practical scenario.

% ==================
\section{Related Works}

Several recent studies have demonstrated the vulnerabilities of GNNs to adversarial attacks \cite{dai2018adversarial,wang2019attacking,wu2019adversarial,zhang2020backdoor,zugner2018adversarial}. These attacks encompass various techniques aimed at deceiving the GNN model, such as altering node features or structural information. Additionally, researchers have explored attacks which aim to steal links from the trained model \cite{he2021stealing,zhang2021graphmi} and extracting private graph information through feature explanations \cite{olatunji2022private}. Property inference attacks have also been launched on GNNs \cite{zhang2022inference}, where an adversary can infer significant graph statistics, reconstruct the structure of a target graph, or identify sub-graphs within a larger graph. Another type of attack, membership inference, distinguishes between inputs the target model encountered during training and those it did not \cite{emmanuelMIA,duddu2020quantifying}. Model inversion attacks aim to infer edges of a graph using auxiliary knowledge about graph attributes or labels \cite{zhang2022model}. The vulnerabilities of GNNs extend beyond these attack techniques, with model extraction attacks \cite{wu2022model} and stealing attacks \cite{shen2022model} posing additional risks. 

Collectively, these studies provide valuable insights into the various vulnerabilities that GNNs are prone to. To the best of our knowledge, no attribute inference attack has been proposed to infer sensitive attributes (node features) from queries to a target model (black-box access). In addition, previous AIAs proposed on tabular data are not applicable to graphs \cite{fredrikson2014privacy,mehnaz2022your,yeom2018privacy,jayaraman2022attribute}. This is because first, all the attacks assume that the data points that are nodes are independent and secondly, they only consider binary attributes. This is not the case for graphs where nodes can be linked to other nodes by some graph properties, the node attributes can be highly correlated, and may have continuous values. Moreover, all other attacks can only infer one sensitive attribute at a time. Here, we propose multi-attribute inference attacks for GNNs. However, it should be noted that \cite{duddu2020quantifying} performed preliminary white-box AIA and their method makes assumptions that the attacker has access to node embedding and sensitive attribute from the auxiliary graph (shadow data). The attacker trains a supervised attack to map embedding to sensitive attribute using the shadow data on the shadow model. The learnt mapping is then used to infer the attribute from the publicly released target embedding. It is important to mention that their approach focuses on the white-box setting where the attacker has access to the internal workings (node embeddings) of the target model and also requires strong assumptions of shadow data and shadow model from the same distribution.

% ==================
\section{Overview}
\mpara{Attribute Inference Attack (AIA).} The task in AIA is to infer sensitive attributes (or features) on a number of nodes. These attributes could hold any possible kind of value. In this paper, we consider datasets having two categories of attribute values, binary and continuous. We note that AIA in its traditional form has only been used for inferring binary attribute values. We take the first step to evaluate on continuous attribute value. We define the task of AIA as follows: 

\begin{definition}[AIA]\label{def:aia}
Let some GNN model $\Phi$ be trained on a graph $\mathcal{G}=(\mathcal{V},\mathcal{E})$ and node feature matrix $\mathbf{X}\in \mathbb{R}^{|\mathcal{V}|\times d}$, let $\mathbf{X}^*\in \mathbb{R}^{|\mathcal{V}|\times d}$ be the partial feature matrix such that the sensitive attributes of a subset of nodes $\mathcal{V}'\subset \mathcal{V}$ are masked/hidden. Given access to $\mathbf{X}^*$ and blackbox access to the GNN model $\Phi$,  the goal of the attacker is to reconstruct the hidden entries of $\mathbf{X}^*$.
\end{definition}

\mpara{Black-box vs White-box access.} Here, we distinguish between two types of model access: black-box and white-box. In a black-box access scenario, an adversary can query the model and receive outputs for the query nodes, but does not have knowledge of the model's internal workings. On the other hand, in a white-box scenario, the attacker possesses knowledge of the target model's internal mechanisms, including its weights and embeddings. However, acquiring such white-box access can be challenging in practice due to intellectual property concerns. Hereafter, we focus on the practical black-box access scenario which is commonly encountered in MLaaS.

\subsection{Threat Model}
\mpara{Adversary's background knowledge.} In AIA, the adversary has access to the trained (target) model, knows the non-sensitive attribute values, and the data distribution (which we later relax). 
We make no assumption about the adversary access to the graph structure but rather experimentally vary such access. Also, the attacker might be interested in inferring multiple attributes. Concisely, we characterize the adversary's background knowledge along two dimensions:\\
\textbf{\textit{Available nodes.}} This quantifies the amount of nodes with sensitive attributes that are available to the attacker. Note, this is different from the sensitive attribute she wants to infer. This refer to \settingone and \settingtwo in our experiments. In \textbf{\settingone}, the attacker only knows the non-sensitive attributes while in \textbf{\settingtwo}, she knows 50\% of the sensitive values and all non-sensitive attributes.\\
\textbf{\textit{Graph structure.}} Whether the attacker has access to the graph structure used in training the target model or not. In the case where she has no access to the graph structure, she artificially creates a KNN graph from the candidate nodes.

% ==================
\section{Proposed Attacks}

Towards this, we develop two different attacks. The first attack which we call \textit{attribute inference attack via repeated query of the target model} (\fpmaattack) involves querying the target model multiple times with attributes produced by a feature propagation algorithm. The attacker iteratively performs feature propagation by querying the target model with the estimated attributes and evaluates the result using a confidence score.
We develop a variant of the attack, \rimaattack, that initializes the sensitive attribute based on a random initialization mechanism and also iteratively queries the target model.

In the second attack which we call \textit{feature propagation-only attribute inference
attack} denoted as \fpattack, relies on a single execution of a feature propagation algorithm. The result obtained from this feature propagation is subsequently considered as the final outcome of the attack. As a baseline, we replace the missing attribute with random values, we refer to this as \riattack attack.
Lastly, we also propose a \textit{shadow-based attribute inference attack} denoted as \saattack that assumes that an attacker has access to a shadow dataset and a corresponding shadow model similar to the target model. We note that this attack has a major limitation in that it may be difficult to obtain such shadow dataset and model in practice but we include it as a relaxed version of our black-box AIA attacks.

Note that all attack methods except \saattack do not utilize any prior information about the labels.

\subsection{Data Partitioning}
For all attacks except \saattack, the dataset is partitioned into train/test sets, which are utilized to train and evaluate the target model. The candidate set is chosen from the training set and includes nodes with missing sensitive attributes for which the attacker wants to infer.

In the case of the \saattack attack, the dataset is initially divided into two parts, namely the target and shadow datasets. The train and test sets are then derived from each dataset to train and evaluate the respective target and shadow models. The candidate set for evaluating the attacks is selected from the training set of the target dataset. All attacks are evaluated using the candidate set \Xst.

\subsection{Attack Modules}
In the following, the major components of the attacks are introduced. Specifically,
feature propagation and confidence score. Feature propagation algorithm is used by the attacks for estimating the sensitive attributes, while confidence score acts as a threshold for measuring the correctness of the inferred attributes.

\mpara{Feature Propagation} 
\label{sec:feature-propagation}
The feature propagation algorithm is a method for constructing missing features by propagating the known features in the graph.
As shown in Algorithm \ref{alg:FPAlg}, the algorithm takes as input a set of nodes~\Xst~which have missing attributes, the adjacency matrix (either given or constructed via KNN), and the number of iterations which is the stopping criteria for the algorithm.

\begin{algorithm}
\LinesNumbered
\DontPrintSemicolon
\SetKwFunction{rand}{AssignRandomAttributes}
\SetKwFunction{known}{AssignKnownAttributes}
\SetKwFunction{glap}{CalculateGraphLaplacian}
\SetKwFunction{round}{Round}
\SetKwInOut{Input}{Input}\SetKwInOut{Output}{Output}
\Input{Partial node feature matrix \textbf{with missing attributes} ~\Xst, adjacency matrix $\mathbf{A}$ and number of iterations~\iter}
\Output{Reconstructed node feature matrix \Xr}
\BlankLine
Compute graph Laplacian \ALap~ = $\mathbf{D}-\mathbf{A}$\ \text{\hspace{0.1cm}where $\mathbf{D}$ is the degree matrix}\;

Assign random attributes to missing attributes in \Xst~\;
\Xr~$\leftarrow$ \Xst\;
\For{~\iter~}{
\Xr~$\leftarrow$ \ALap\Xr \text{\hspace{5em}$\backslash\backslash$Propagate attributes}\;
\If{attributes values are binary}{
\Xr~$\leftarrow$ \round{\Xr} \text{\hspace{2em}$\backslash\backslash$ round to 1 if attributes $\ge$ 0.5, else to 0}\;
}
\Xr~$\leftarrow$ \known{\Xst}
}
\KwRet{\Xr}
\BlankLine
\caption{Feature propagation algorithm}
\label{alg:FPAlg}
\end{algorithm}
Feature propagation starts by computing a normalized Laplacian of the graph~\ALap~(line 1), then initializes only the missing attributes in~\Xst~with random values. This randomly initialized~\Xst~is denoted as~\Xr~(line 3).~\Xr~is propagated over the graph by multiplying it with the computed graph Laplacian~\ALap~(line 4). This step is repeated for multiple iterations until convergence \cite{rossi2021unreasonable}. Since convergence might be difficult to attain on a real-world dataset, we fix the values of iterations experimentally. If the attribute values are binary, the updated values of~\Xr~are rounded up such that any values above $0.5$ are assigned the value 1 and 0 otherwise (line 7). For a dataset with continuous values, this step is omitted. Then, the values of the known attributes in~\Xst~(attributes that were not considered missing at the start of feature propagation) will be reassigned to nodes in~\Xr~(line 9). Since feature propagation is an algorithm that reconstructs missing values by diffusion from the known neighbors, the non-missing attributes of the neighbors are always reset to their true values after each iteration.

\mpara{Confidence Score} 
\label{sec:confidence-score}
In our attacks, since the attacker has no information about the ground truth label, she utilizes the confidence of the model based on its prediction. The underlying idea is that a model that is confident in its prediction will assign a higher probability to the corresponding class label. Thus, the attacker leverages this confidence as a measure of the confidence score. One approach is to consider the value of the class with the highest probability as the confidence score. However, a problem arises when the target model generates an output where either all classes have similar probabilities or there are multiple classes with comparable probabilities.

To address this issue, we propose a solution by applying a ``tax'' on the highest class probability.
First, we compute the average class probability of the remaining classes, and then determine the difference between this average and the highest probability in $\mathbf{y}$.
This difference serves as the confidence score.
Intuitively, if the class probabilities in $\mathbf{y}$ are similar, the final score will be low, indicating a lower level of confidence. Conversely, if there is a substantial difference between the highest class probability and the others, the taxation is reduced, resulting in a higher final confidence score.
It is important to note that the output vector $\mathbf{y}$ is normalized, ensuring that the maximum confidence score is 1 and the minimum is 0. Additionally, the confidence score is computed on a per-node basis.

\subsection{Attribute Inference Attack via Repeated Query of the Target Model (FP-MA)}
\begin{figure}[htbp] %[h]
\centering
\includegraphics[scale=.32]{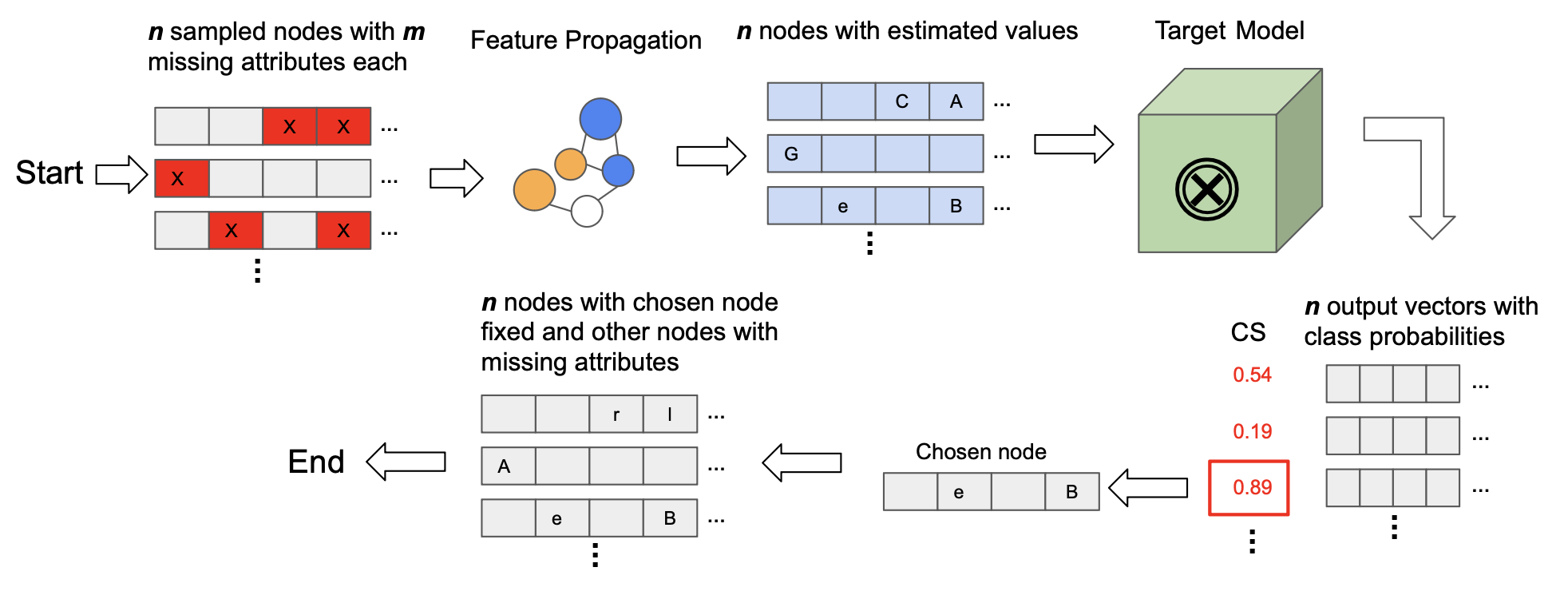}
\caption{The black-box attack \fpmaattack. For the \rimaattack attack, we replace the feature propagation module with a random initializer.}
\label{pic:AIAAttack}
\end{figure}

Our first attack, referred to as \fpmaattack, involves multiple queries to the target model in order to infer the sensitive attribute. It relies on the feature propagation algorithm (Algorithm \ref{alg:FPAlg}) to initialize the sensitive attribute. The attack process is depicted in Figure \ref{pic:AIAAttack}. Furthermore, we introduce another variation of the attack, known as \rimaattack, which initializes the sensitive attribute through a random initialization mechanism while accessing the target model. The overall procedure of the attack is outlined in Algorithm \ref{alg:MainAttackAlg}.

\begin{algorithm}[htbp] %[h!]
\DontPrintSemicolon
\LinesNumbered
\SetKwData{i}{$i$}
\SetKwFunction{Init}{InitAlgorithm}
\SetKwFunction{gcn}{TargetGCN}
\SetKwFunction{cs}{CalculateConfidenceScores}
\SetKwFunction{max}{IndexOfMax}
\SetKwFunction{lowerby}{LowerBy5Percent}
\SetKwFunction{meanofcs}{Mean}
\SetKwInOut{Input}{input}\SetKwInOut{Output}{output}
\Input{Incomplete node feature matrix with missing sensitive attributes ~\Xst, Edges \ei, confidence threshold~\CSth}
\Output{\Xr~with inferred attributes, mean confidence score~\CSmean}
\BlankLine
\Xr~$\leftarrow$\Xst\;
\BlankLine

\While{~\Xr~has missing values}{
\Xr~$\leftarrow$\Init{\Xr, \ei} \text{\hspace{0.7em}$\backslash\backslash$Do Feature Propagation}\;
\Y~$\leftarrow$\gcn{\Xr,\ei}\text{\hspace{0.7em}$\backslash\backslash$Query target Model}\;
\CS~$\leftarrow$\cs{\Y}\;
\eIf{~\CS~has a value greater than~\CSth~}{
\i$\leftarrow$\max{\CS} \text{ \hspace{0.2em}$\backslash\backslash$ Check index of maximum Confidence Score (\CSsmall)}\;
\Xr$[\i]~\leftarrow $max(\Xr$[\i])_ \text{\CSsmall}$\text{ \hspace{0.2em}$\backslash\backslash$ Choose node with maximum~\CSsmall}\;
\text{Reset}~\CSth\;}{
\CSth~$\leftarrow$ \lowerby{\CSth}\;
}
}
\BlankLine
\Y~$\leftarrow$\gcn{\Xr, \ei}\;
\CS~$\leftarrow$\cs{\Y}\;
\CSmean~$\leftarrow$\meanofcs{\CS}\;
\BlankLine
\KwRet{\Xr,~\CSmean}
\BlankLine
\caption{Attack algorithm. \texttt{InitAlgorithm} refers to either the feature propagation (FP-MA) or the random initilization (RI-MA)}
\label{alg:MainAttackAlg}
\end{algorithm}

First, the attacker has~\nsamp~candidate nodes with~\mmissing~missing sensitive attributes, the corresponding edges of these nodes (or she computes the edges via KNN if she has no access), and a confidence score threshold~\CSth. These nodes are in the matrix~\Xst. In this first phase, the attack procedure runs the initialization algorithm (feature propagation (Section \ref{sec:feature-propagation}) or random initialization) to obtain an estimated value for the missing attribute (line 3). The attacker then queries the target model with the attributes obtained from the initialization algorithm \Xr~and the edges \ei~(line 4). As shown in lines 5-7, the attacker computes the confidence scores as described in Section~\ref{sec:confidence-score}~and then chooses the node which produces the highest confidence score if it passes the confidence score threshold. Any node whose confidence score is higher than the threshold is ``fixed'' (line 8). That is, the estimated values for the missing attributes of those nodes does not change in the next iteration of the attack.

To incentivize the attack to infer nodes with high confidence scores, a threshold for the confidence score~\CSth~is selected by the attacker at the start. A node with inferred attributes with a confidence score lower than the threshold will not be fixed (line 11). The threshold assures that the algorithm will have multiple iterations to maximize the confidence score and in turn predict the right attribute value. But a problem arise when the threshold is too high and no nodes can obtain such confidence score. To tackle this problem, we propose a method that lowers the confidence score by 5\% after each iteration when a node is not fixed (line 11). The lowered threshold is set back to the original value when a node is finally fixed (line 9). We reset the threshold to ensure that the attacker is re-incentivized when the feature propagation algorithm produces new randomized values for the rest of the nodes. This is an iterative process and are repeated until no nodes are left with missing values.

When the attacker infers values for all nodes with missing sensitive attributes, it queries the target model with these nodes and edges, compute the confidence scores, and then takes the mean of all the confidence scores of all inferred nodes (lines 14-16). The mean of the confidence score is returned for experimental purposes to compare the behavior of the confidence score to other attack methods. The attacker finally returns the candidate nodes with their inferred attributes~\Xr~and the mean of the confidence scores~\CSmean~(line 17).

\subsection{Feature Propagation-only Attribute Inference Attack (FP)}

The feature propagation-only attribute inference
attack (\fpattack) only execute a feature propagation algorithm once to infer the sensitive attribute.

One advantage of \fpattack is that it is simpler and faster in runtime than other methods. This is because \fpattack attack does not utilize the information obtained by querying the model in the attack process. 
The attack procedure is as follows. First, \fpattack takes the missing attributes in~\Xst~and their edges \ei~as input. The attacker then runs feature propagation algorithm as described in Section \ref{sec:feature-propagation}.
% ~(line 1). 
The output of the feature propagation algorithm is considered as the inferred nodes.
% (lines 2). 
Similar to \fpmaattack, the target model is queried with the final inferred nodes, not for finetuning the inferred estimate but only to compute the mean of the confidence scores as a measure of comparison with other methods.
% (lines 3-5). 
Finally, \fpattack returns the candidate nodes with their inferred attributes \Xr~and the mean confidence~\CSmean.
% (line 6).

\subsection{Shadow-based attribute inference attack (SA)}
We adapt (with several modification) the white-box attack proposed by \cite{duddu2020quantifying} into the black-box setting. In this attack, the model's output (posteriors) is used to determine whether the attribute has been correctly inferred or not. The purpose of the attack is to study the behavior of the model on nodes with sensitive attribute values that have already been seen. To achieve this, a pseudo-model called a shadow model is trained, which the attacker can use to observe the model's behavior on her shadow dataset. The attacker chooses a candidate set from the train set of her shadow model, which includes nodes with complete attributes and some with missing sensitive attribute values. During the query, she randomly assigns random values to the missing sensitive attributes, observes the posterior, and compares it with the posterior of the node-set with the true attributes. 
After obtaining the labeled posteriors from the candidate set of the shadow dataset, she proceeds to train the attack model, which is a 3-layer MLP model. In this step, the attack model is trained using the labeled posteriors, where the posteriors with the original sensitive attribute value are labeled as $1$, while those with the assigned random attribute value are labeled as $0$. To infer the attribute of the candidate set (node of interest), she queries the target model, obtains posteriors, and inputs them into her trained attack model to infer attribute values. We note that this attack is only applicable for sensitive binary attribute values.

% ==================
\section{Datasets and Implementation details}
% \textbf{Datasets}
We utilize three private datasets, namely the credit defaulter graph, Facebook, and LastFM social network dataset, along with two benchmark datasets, Cora and PubMed. The details for all the datasets are provided in Table \ref{tab:datasets-table} and in Appendix \ref{sec:datasets-desc}. The target GNN model $\Phi$ is a 2-layer graph convolution network (GCN). All our experiments were conducted for 10 different instantiations. We report the mean values across the runs in the main paper. The full result with standard deviation is on GitHub. \\

\begin{table}[htbp] %[h]
\caption{Statistics of dataset used in all experiments}
\centering
\footnotesize
\begin{tabular}{lc@{\hspace{0.2cm}}c@{\hspace{0.2cm}}c@{\hspace{0.2cm}}c@{\hspace{0.2cm}}c@{\hspace{0.2cm}}c}
\toprule
& Credit & Cora & Pubmed & Facebook & LastFM & Texas \\\midrule
\# attributes & 13 & 1,433 & 500 & 1,406 & 128 & 10 \\
$\mathcal{|V|}$ & 30,000 & 2,708 & 19,717 & 4,167 & 7,624 & 925,128\\
$\mathcal{|E|}$ & 1,436,858 & 5,278 & 44,324 & 178,124 & 55,612 & --\\
$\mathnormal{deg}$ & 95.79 & 3.90 & 4.50 & 42.7 & 7.3 & -- \\
\# classes & 2 & 7 & 3 & 2 & 18 & 100\\
Train Acc. & 0.78 & 0.95 & 0.88 & 0.98 & 0.80 & 0.53 \\
Test Acc. & 0.77 & 0.78 & 0.86 & 0.98 & 0.85 & 0.45 \\
\bottomrule
\end{tabular}
\label{tab:datasets-table}
\end{table}

% ==================
\mpara{Attack Evaluation}
For all experiment, we choose $100$ candidate nodes (\Xst) at random from the training set with the objective of inferring the sensitive attribute values. Note that these candidate nodes are fixed across all experiments unless otherwise stated. To assess the success of the attack, we compare the inferred attributes with the values of the original node-set, which refers to the nodes that were not modified. We employed two metrics that measures the percentage of the inferred attributes for binary values and mean-squared error for continuous values. Details of the evaluation methods are in Appendix \ref{sec:evaluation_metric}.

\section{Results}

\begin{figure}[htbp] %[h]
\centering
\includegraphics[width=1\linewidth]{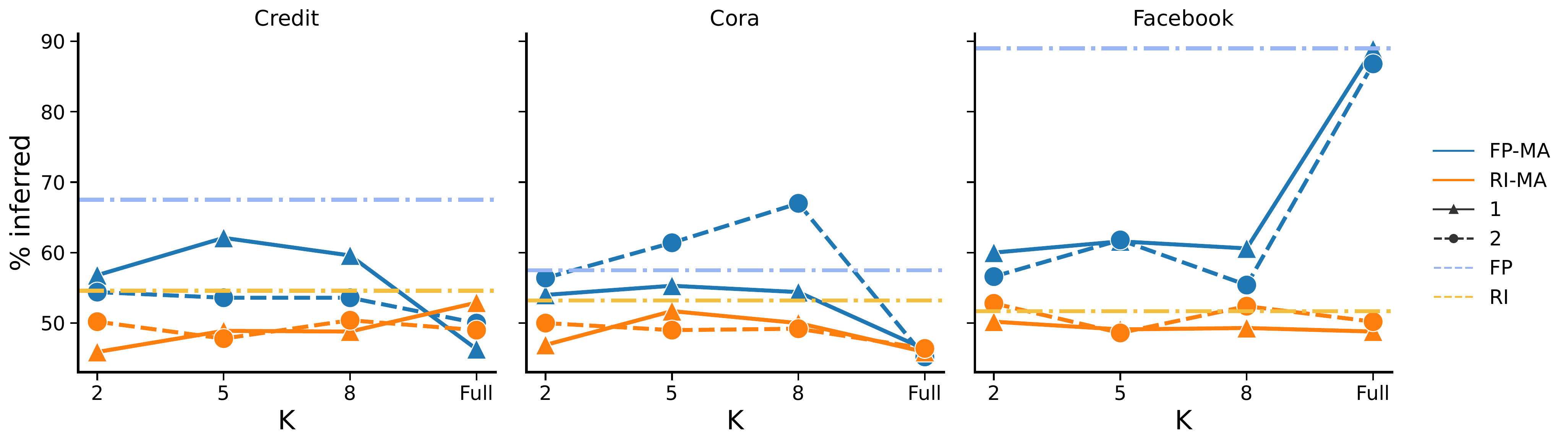}
\caption{Performance of the binary attribute inference attack at different settings. \settingone is represented by a solid line with triangle marker, while \settingtwo is represented by a dashed line with a circle marker. We fix the size of the training dataset to 1000 for all datasets.}
\label{fig:fp_ri_influence_crecorfac}
\end{figure}

\mpara{Inference of sensitive binary attributes.}
In the case when the missing sensitive attributes are binary, we observe a similar trend among all datasets: the attack performance is worse when the attacker has access to a black-box model compared to when they have no access, as shown in Figure \ref{fig:fp_ri_influence_crecorfac}.
In \settingone, \fpattack and \riattack, which do not require access to the target model, exhibit slightly better performance (an improvement of at most 4\%) compared to the black-box attack models \fpmaattack and \rimaattack, which rely on such access. It is important to note that in \settingone, the attacker only has information on all non-sensitive nodes and no additional knowledge about some of the sensitive attributes, unlike in \settingtwo. When \fpattack and \riattack outperform \fpmaattack and \rimaattack respectively, it suggests that even with access to the trained model, no further sensitive information can be leaked. One reason for this is that the available information from non-sensitive nodes already captures the majority of patterns and correlations in the dataset. Therefore, the incremental advantage gained from accessing the model is minimal, resulting in comparable or slightly inferior performance of the black-box attack methods. In \settingtwo, the performance of \riattack is similar to \settingone, where \riattack performs slightly better than \rimaattack in inferring the sensitive attribute. However, the opposite phenomenon is observed for \fpattack. Specifically, \fpmaattack achieves a higher inferred attribute rate. For instance on the Cora dataset, this difference amounts to an improvement of 21\% .

On all datasets except Facebook, we observe that when the attacker has access to the graph structure, it does not provide additional advantages to the attack models (\fpmaattack and \rimaattack). Computing the graph based on K nearest neighbor strategy to query the target model leads to better inference than access to the true graph structure. One reason for this is that the candidate nodes are disjoint from the training nodes, and the connections among the candidate nodes are relatively sparse. Therefore, when using the original sparse neighborhood the feature representation of the query node might be less informative as compared to using the neighborhood constructed based on feature closeness.

For the \saattack attack, the attack performance is a random guess ($<=50\%$) on all datasets. Therefore, in addition to the difficulty in obtaining a shadow dataset, such attacks are also not successful given a black-box access to the model. We omit the results due to space limitation.

\begin{figure}[htbp] %[h]
\centering
\includegraphics[width=0.8\linewidth]{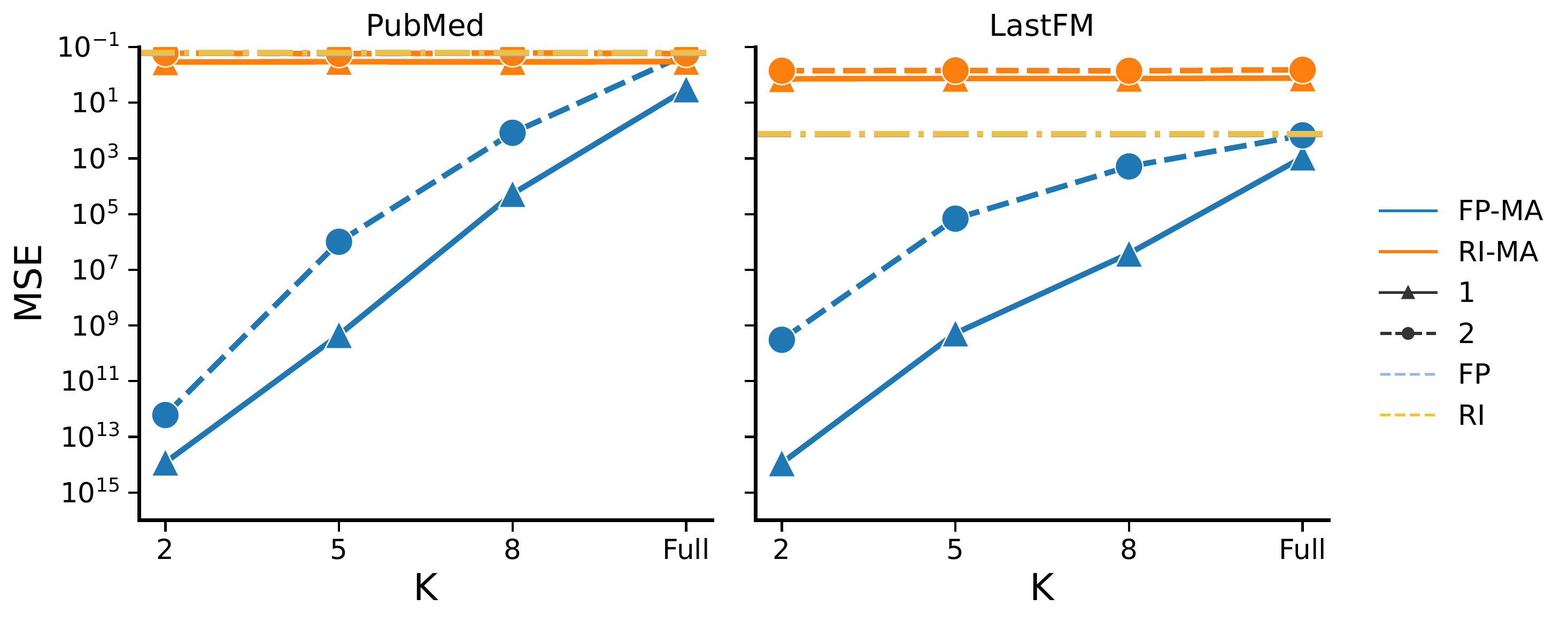}
\caption{Performance of inferring continuous attributes on the PubMed and LastFM dataset. \settingone is represented by a solid line with triangle marker, while \settingtwo is represented by a dashed line with a circle marker. We fix the size of the training dataset to 1000.}
\label{fig:fp_ri_influence_publast}
\end{figure}

\mpara{Inference of sensitive continuous attributes.}
As shown in Figure \ref{fig:fp_ri_influence_publast}, on the LastFM dataset, \fpmaattack performs the worst, while \rimaattack achieves the best results. The performance of \fpattack varies depending on the availability of the full graph structure or using the KNN approach. 
Throughout the experiments, using the full graph structure consistently yields better results for \fpmaattack. The \riattack and \rimaattack method, on the other hand, is not sensitive to the graph structure. \rimaattack method consistently outperforms the other methods in \settingone.

In \settingtwo, the performance pattern follows a similar trend as \settingone, but with improved results. This can be attributed to the attacker having knowledge about some sensitive attributes. For example, at 100 training dataset size, the error rate drops by 60\% for \fpattack, 45\% for \fpmaattack, 51\% for \riattack, and 50\% for \rimaattack. This highlights the impact of having knowledge of nodes with sensitive attributes. An attacker with such information can launch a stronger attack compared to an attacker with no knowledge, as the error rates significantly decrease when partial knowledge is available.

Similarly, on the PubMed dataset (%Table \ref{tab:pubmed-single-attrib} and 
Figure \ref{fig:fp_ri_influence_publast}), we observe that an attacker in \settingtwo can infer the sensitive attributes better compared to an attacker in \settingone, with a significant decrease in error rates of up to 52\%.
Among the inference methods used, \fpmaattack consistently achieves the best performance (lowest error) in inferring the sensitive attribute across all settings in the PubMed dataset. The reason behind this superiority can be attributed to the utilization of feature propagation and the availability of the graph structure. By leveraging feature propagation algorithms, \fpmaattack can effectively propagate information and exploit the relationships between nodes to make more accurate inferences. 

In both the settings, having access to the graph structure consistently yields the highest inference capability (lowest error rates) for all other methods except for \riattack and \rimaattack. This observation is not surprising unlike in the case of inferring binary attributes. This is because continuous attributes typically exhibit a smooth and gradual variation, whereas binary inference focuses on identifying distinct decision boundaries within the data. Hence, the connectivity patterns of the nodes in the graph structure play a crucial role in propagating information and inferring missing values. The inference methods leverage these patterns to make accurate estimations.

\textbf{Summary.} 
For inferring binary attribute, the underlying algorithms used by \fpattack and \riattack methods allow them to leverage the dataset's inherent characteristics effectively. On the other hand, \fpmaattack and \rimaattack, relying on access to the target model, may exhibit slightly lower performance in the absence of specific knowledge about the sensitive attributes such as in \settingtwo. Additionally, successful attacks can be carried out without full access to the graph structure.

For inferring continuous attribute, the results emphasize the importance of utilizing the graph structure, and additional information in improving the accuracy of inferring the sensitive attributes.

\begin{figure}[h] %[h]
\centering
\includegraphics[width=1\linewidth]{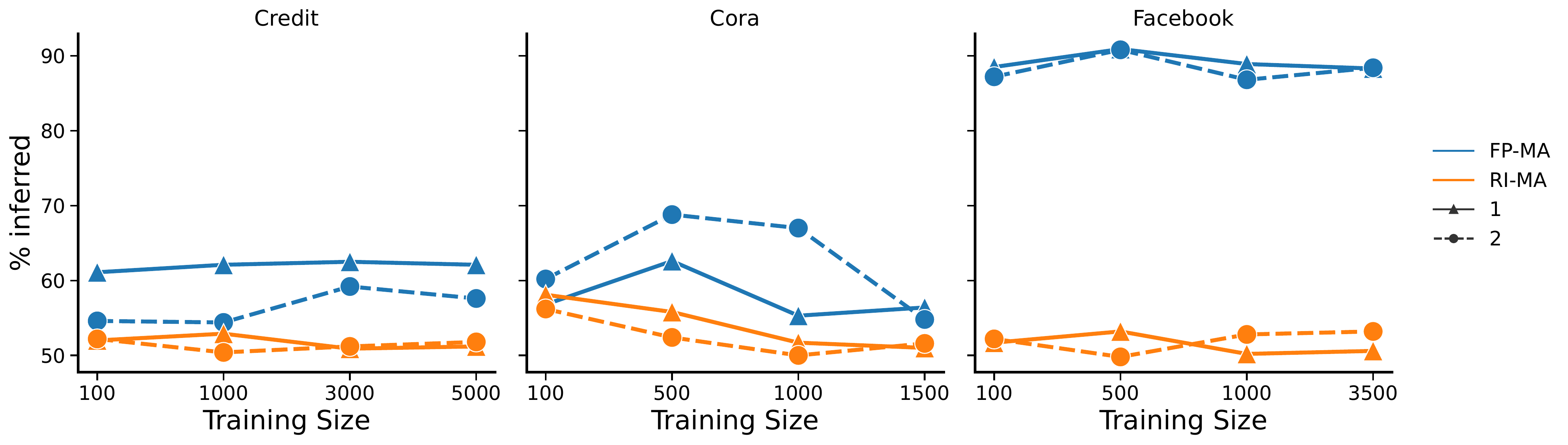}
\caption{Performance of varying target model's training size on black-box attacks (\rimaattack and \fpmaattack) on binary AIA.}
\label{fig:crecorfacrima_fpma}
\end{figure}

\mpara{Effect of the Training Data Size on Inferring Sensitive Attributes}
\label{sec:effect-training-size}
Here, we investigate the influence of the data size used to train the target model on the inference of the sensitive attribute.
On the Cora and Facebook dataset (Figure \ref{fig:crecorfacrima_fpma}), we observe that it is easier to infer more attributes when there is less data used in training the target model. One reason for this is that when there is less training data available, the target model may not have learned all the relevant features, and therefore the attack can leverage this to infer the sensitive attribute more easily. However, as the number of available training data increases, the target model becomes better at learning the underlying patterns in the data, which in turn makes it more difficult for an attacker to infer sensitive attributes from the model's output. Additionally, the increased training data may also reduce overfitting and improve generalization performance, which can make the model less vulnerable to attacks. We also observe that the attack achieves greater success in \settingtwo compared to \settingone. This outcome is expected, as it demonstrates the attack's ability to leverage the information from the known 50\% sensitive attribute values to enhance the accuracy of inferring the remaining sensitive attribute values.

\begin{figure}[h] %[h]
\centering
\includegraphics[width=0.8\linewidth]{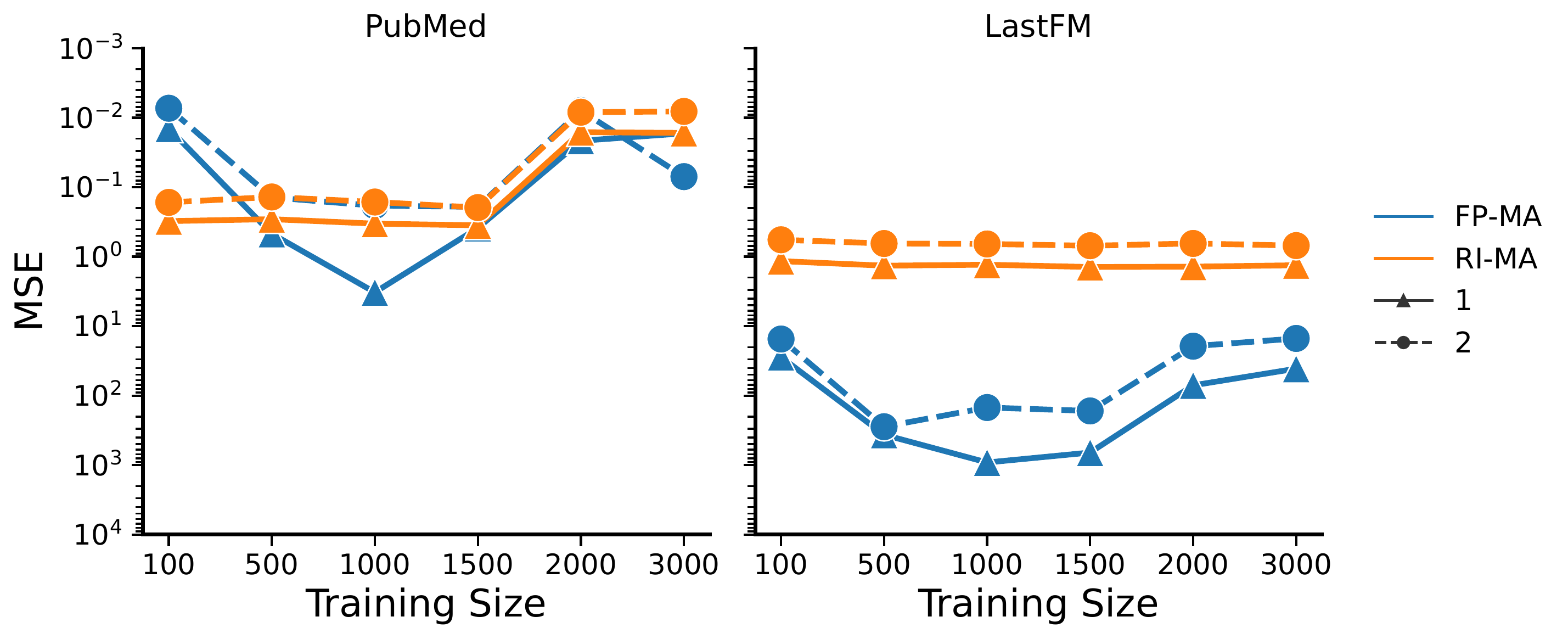}
\caption{Performance of varying target model's training size on black-box attacks (\rimaattack and \fpmaattack) on continuous AIA}
\label{fig:publastrima_fpma}
\end{figure}

On the Credit dataset in Figure \ref{fig:crecorfacrima_fpma}, the effect of the target's training data on inferring the sensitive attribute is minimal. For instance, the performance of \fpmaattack is similar across different variations of training size.
Furthermore, the additional knowledge of certain sensitive attributes (\settingtwo) does not have any noticeable effect.
On the Pubmed and LastFM datasets (Figure \ref{fig:publastrima_fpma}), \rimaattack consistently achieves lower error rates as training data sizes increases. This indicates that \rimaattack benefits from larger training datasets in terms of capturing more diverse patterns leading to improved performance and lower error rates.
However, for \fpmaattack, we observed a convex-shaped effect as the training data size increases. Although this is strange, we believe that as the training size increases, the model may encounter more outliers or noise that hinder its performance, resulting in higher error rates. However, as the training size further increases, the model starts to learn the underlying patterns better, leading to improved performance and lower error rates.

\begin{figure}[h!] %[htbp]
\centering
\includegraphics[width=1\linewidth]{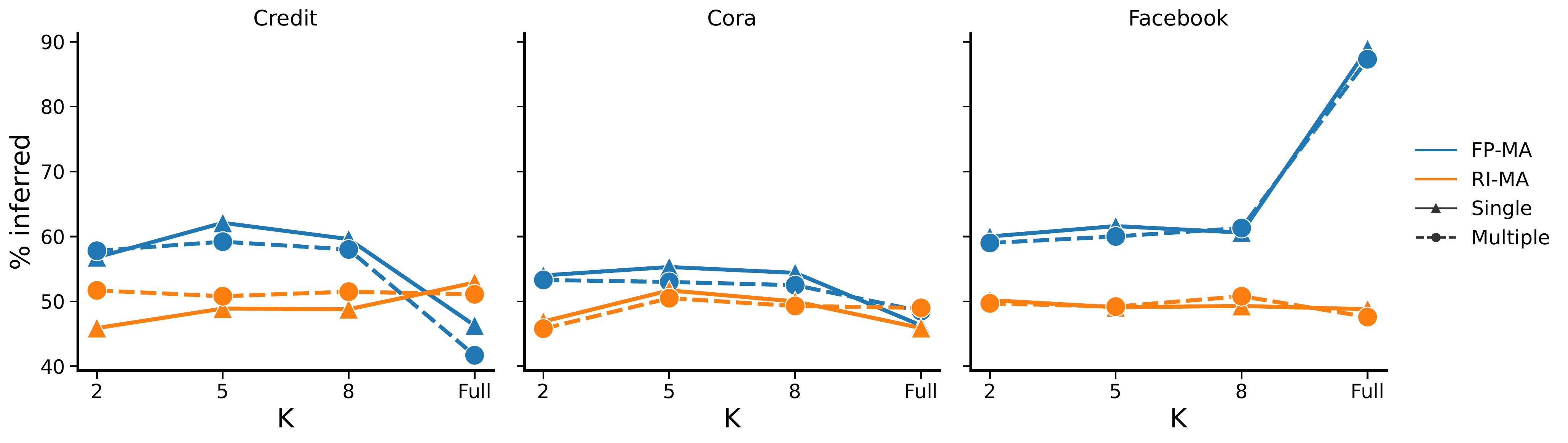}
\caption{Performance comparison of single and multiple attribute inference attack on the Credit, Cora, Facebook dataset}
\label{fig:mul_vs_sin_crecorfac}
\end{figure}

\mpara{Inferring Multiple Attributes (m$>1$)}
\label{sec:multiple-attributes}
In the multiple attributes inference experiment, the attacker is interested in inferring more than one attribute. For example, on the Credit dataset, the attacker may want to infer both the age and education level of the victims. In this experiment, we set $m=2$. As shown in Figure \ref{fig:mul_vs_sin_crecorfac}, the results of the multiple attribute inference closely follow the trends observed in the case of single attribute inference. However, one notable difference is that on the Credit, Cora, and Facebook datasets, the performance of inferring multiple attributes is lower as compared to single attribute inference (solid lines). This is expected because of the increased complexity to the inference task. The presence of multiple attributes introduces additional dependencies and interactions among the attributes, making it more challenging to accurately infer all attributes simultaneously. Moreover, some attributes may have conflicting patterns or dependencies, making it difficult for the inference algorithm to reconcile the competing information. 

\begin{figure}[h!] %[htbp]
\centering
\includegraphics[width=0.8\linewidth]{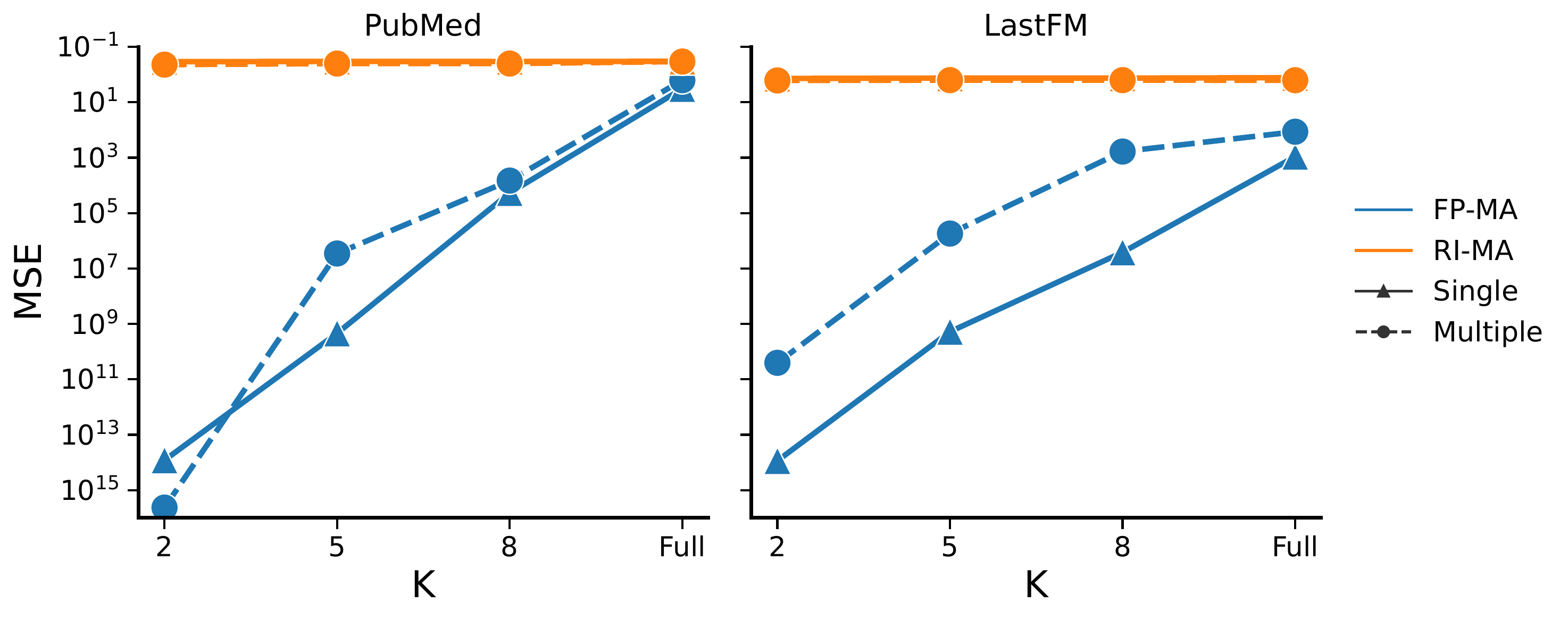}
\caption{Performance comparison of single and multiple attribute inference attack on the Pubmed and LastFM dataset}
\label{fig:mul_vs_sin_publast}
\end{figure}

For inferring sensitive continuous values, the opposite is observed, especially for \fpmaattack (Figure \ref{fig:mul_vs_sin_publast}). Specifically, inferring multiple attributes achieves lower error rates than inferring a single attribute, with up to a 99\% decrease in error on both the PubMed and LastFM datasets (dashed lines). This interesting phenomenon can be attributed to the unique characteristics of the feature propagation algorithm used in \fpmaattack. The feature propagation algorithm utilizes the relationships and dependencies among the attributes to propagate information and refine the inferred values. When inferring multiple attributes simultaneously, the propagated information from one attribute can provide valuable insights and constraints for the inference of other related attributes. Specifically, if certain attributes have missing or noisy data, the presence of other attributes with similar patterns may compensate for the errors and improve the robustness of the inference process.

\mpara{Additional Experiment} We perform additional experiment by varying the distribution assumption in Appendix \ref{sec:aia-tabular-data}. The result on the Texas dataset demonstrates that having access to skewed distributions (where the candidate and the training nodes are from different distributions) leaks more information than having access to the same distribution when the training dataset size is small.

\section{Conclusion}
In this paper, we develop several black-box attribute inference attacks to quantify the privacy leakage of GNNs. Our findings are as follows: \\
(i) For a stronger attacker with additional access to some sensitive attribute (\settingtwo), the performance of black-box attacks can improve by up to 21\% compared to an attacker without such access (\settingone). \\
(ii) The graph structure plays a significant role in inferring sensitive continuous values, leading to a remarkable reduction in error rate of up to 99\%. However, when it comes to inferring sensitive binary values, except for the Facebook dataset, the graph structure has no noticeable impact. \\
(iii) Despite a stronger attacker (\settingtwo) and access to the graph structure, our black-box attribute inference attacks generally does not leak any additional information compared to missing value estimation algorithms, regardless of whether the sensitive values are binary or continuous.\\ \\
% %-------------------------------------------------------------------------------
{\small \textbf{Acknowledgment.}
This work is, in part, funded by the Lower Saxony Ministry of Science and Culture under grant no. ZN3491 within the Lower Saxony ``Vorab'' of the Volkswagen Foundation and supported by the Center for Digital Innovations (ZDIN), and the Federal Ministry of Education and Research (BMBF), Germany, under the project LeibnizKILabor (grant no. 01DD20003).}

\bibliographystyle{plain}
\bibliography{main}

\appendix
\section*{Appendix}
\section{Performance of varying data the distribution assumption}
\label{sec:aia-tabular-data}
We perform additional experiment to vary the data distribution assumption. For this experiment, we adopted the Texas dataset from \cite{jayaraman2022attribute} and their corresponding training distribution and data availability assumptions. In this case, we used KNN to construct edges among the data with reference values $k=\{2, 5, 8\}$ and report the performance of the best performing $k$. In \cite{jayaraman2022attribute}, three training distribution assumptions were considered; same data distribution and two skewed distributions. For the skewed distribution, it implies that the candidate set is comprised of uniformly sampled data across all hospitals rather than from the training set.
The first skewed distribution ($D_{LP}$) consist of set of $266$ hospitals that have the lowest population of patients, while the second skewed distribution ($D_{HP}$) consists of set of $7$ hospitals that have the highest population of patients.
The data availability assumption refers to the amount of data used in training the target model. For the data availability, we used the reference values $\{100, 50\text{K}\}$ where 50K implies 50,000.

\begin{figure*}[htbp] %[h]
\centering
\includegraphics[width=1\linewidth]{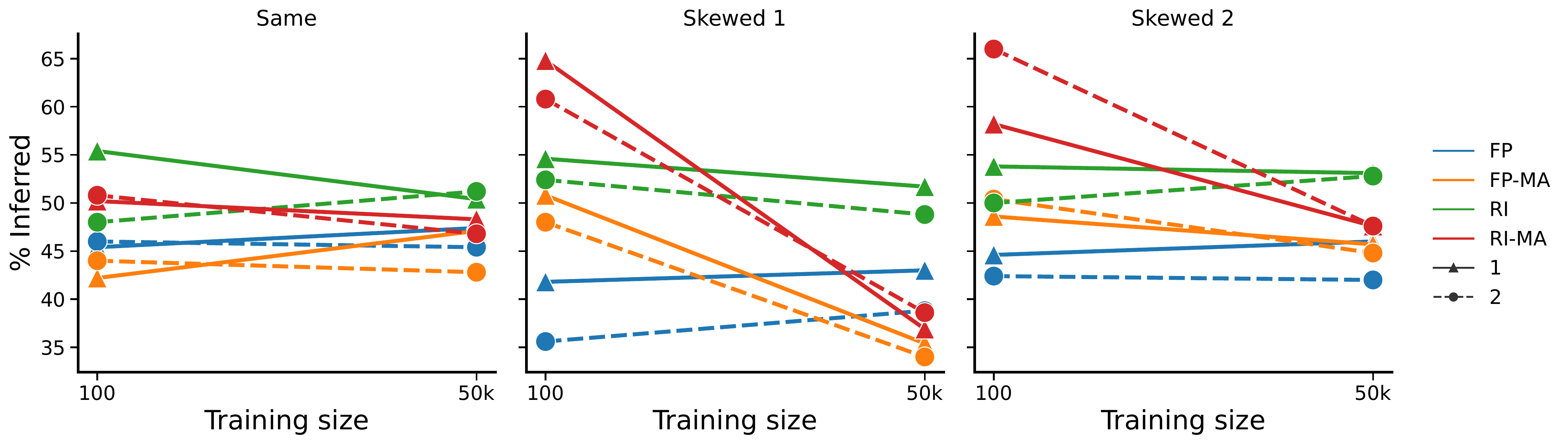}
\caption{Performance comparison of \fpattack, \fpmaattack, \riattack and \rimaattack on the Texas dataset. \settingone is represented by a triangle marker and solid line, while \settingtwo is represented by a circle marker and dashed line.}
\label{fig:texasfp_fpma_ri_rima}
\end{figure*}

As shown in Figure \ref{fig:texasfp_fpma_ri_rima}, overall, the result show that inference attacks based on random initialization (green and red) performs better than any feature propagation based attacks (orange and blue) in all setting. One reason is because the edges constructed among data points based on KNN might not be reflective of the true connectivity among the data points which further confuses the attack model. Furthermore, the result indicates that using the output of the target model can significantly improve the performance of attribute inference attacks, and random initialization can also be a useful method for inferring missing attributes in cases where the graph information is not available or informative. Finally, we observe that having more data available for training does not lead to better inference. One possible reason is the presence of noise or irrelevant information in the larger dataset. As the size of the training dataset increases, it is likely that the dataset becomes more diverse, containing a wider range of attribute values and patterns. However, this increased diversity may also introduce noise or irrelevant information that can hinder the accuracy of attribute inference.

\mpara{Influence of training distribution and size on attack performance}
In the following, we compare the influence of the of the data distribution and data size on the attack performance. Overall, we observe that the attacker with access to the same distribution can infer less sensitive attribute than attacker with a skewed distribution. One reason for this is the potential presence of unique or rare attribute patterns in the skewed distribution (candidate set coming from a different distribution than training set). In such cases, the skewed distribution may contain certain attribute patterns or combinations that are more prevalent or distinct within specific subgroups of the hospitals. These unique patterns, which may not be as prominent in the same distribution, can provide valuable information for the attacker to infer sensitive attributes. Hence, the attacker can exploit these biases to make more accurate inferences about the sensitive attributes.

\mpara{Same data distribution}
In the setting where the attacker has access to the data from the same training distribution, the sensitive  attribute can be inferred by random initialization (\riattack) without the need for access to the target model. This leads to 10\% more than \rimaattack. However, in \settingtwo, \rimaattack outperforms all other inference methods. This can be attributed to the knowledge of partial sensitive records for which the attack method utilize. In all, \fpattack and \fpmaattack infers the least attribute values.

We observe that it is easier to infer attributes when less data are used for training in the case that the attacker has access to data from the same distribution. We believe that this is due to the limitations in data representativeness when data is large.
Additionally, the value of \textit{k} does influences the inference performance with \textit{k=2} the best performing on all settings and training data sizes. Overall, the results suggests that the attack performance is not affected significantly by the size of the training set or the similarity between individuals in the training set.

\mpara{Skewed data distribution 1 ($D_{LP}$)}
In the first skewed data setting which considers hospitals with the lowest number of patients, we observe that \rimaattack outperforms all other methods in inferring the sensitive attributes on all settings and data sizes. This is because the random initialization allows the inference process to explore a wider range of possible attribute values, enabling the attacker to capture rare / distinct attribute patterns or outliers that may be more prevalent in the hospitals with the lowest number of patients. Also, the value of the \textit{k} parameter can have a significant impact on its performance. For instance, the attack performance at \textit{k=2} is 63.4 while at \textit{k=5} is 64.8. This inference performance plunges with higher \textit{k} (\textit{k=8}) which achieves a performance of 62.6. This indicates that the inference performance in the skewed distribution ($D_{LP}$) is sensitive to the value of \textit{k}. Similar to the result when the attacker has access to the same distribution, the attack performance does not increase with an increase in the training data. In \settingone, where the attacker has no access to the sensitive attribute, and \settingtwo, where the attacker has partial access to the sensitive attribute, we observe that the best performance is achieved in \settingone, despite the lack of access to the sensitive attribute with success rates of 64.8\%. We attribute the reason for such performance to the model bias. Specifically, the partial access to the sensitive attribute may inadvertently influence the attacker's model and bias the inference process towards the available partial information. This bias can lead to inaccurate inferences about the missing portion of the sensitive attribute. In contrast, in \settingone, where the attacker has no access to the sensitive attribute, the inference process remains unbiased, allowing the attacker to explore the available attributes more comprehensively. In addition, \fpattack and \fpmaattack are the least methods for inferring sensitive attributes similar to observation when data from the same distribution is available to the attacker.

\mpara{Skewed data distribution 2 ($D_{HP}$)}
Similar to the result of ($D_{LP}$), among all attack methods, \rimaattack achieved the best sensitive inference value with the success rates of 66\%. We observe that an attacker with knowledge of some sensitive value (\settingtwo) has more advantage than an attacker with no knowledge of any sensitive attribute. In particular, up to 14\% increase in the percentage of inferred attribute. This is contrary to the results of ($D_{LP}$). The difference in performance on different settings could be due to the variation in data distribution and the complexity of the target model. The result implies that the attacker can utilize the known attributes as prior to achieve better inference performance since the data distribution comes from hospitals with highest number of patients. 
The choice of $k$ value also affects the performance of the attack methods, where a higher $k$ value means more neighbors are considered, which increases the computation time but can improve the performance.
We observe up to 27\% increase when $k=5$.
Similar to ($D_{LP}$), the attack methods that are based on feature propagation algorithm are the least performing methods for inferring sensitive attributes.

\section{Datasets}
\label{sec:datasets-desc}
\textbf{Credit defaulter graph \cite{agarwal2021towards}.} The credit graph consists of interconnected nodes representing individuals, whose spending and payment patterns were used to establish connections. The objective is to predict whether a person will default on their credit card payment or not, while taking into account their age as the sensitive attribute. \\
\textbf{Facebook \cite{leskovec2012learning}.} The Facebook dataset comprises of nodes representing different user accounts on the social network. Each user node has different features
including the gender, education, hometown. We take the gender as the sensitive attribute. \\
\textbf{LastFM \cite{rozemberczki2020characteristic}.} The LastFM dataset represents mutual follower relationships among users from the social media platform. Each user possesses certain attributes, including preferred music, artists, and location. This is a dataset with continuous attribute values. \\
\textbf{Conventional datasets.} We conducted further tests on commonly used benchmark datasets, Cora and Pubmed. These datasets are citation networks where each node represents a document, and each edge represents a citation. The attributes of each node are represented by a bag-of-words vector, and the labels indicate the category of the document. The attribute values of Cora are binary, while in Pubmed, they are continuous.

\section{Evaluation metric}
\label{sec:evaluation_metric}
\paragraph{Evaluation metric for attacks on binary attributes.} 
To evaluate attacks on datasets with binary attribute values, we measure the percentage of correctly inferred attributes. This is done by computing the \textit{hamming distance} between the node set with inferred attributes and the original node attributes.
If the values of both original and inferred attributes are identical, the hamming distance is $0$, and if not, it is $1$. A lower hamming distance corresponds to a higher percentage of inferred attributes, indicating a more successful attack. 

\paragraph{Evaluation metric for attacks on continuous attributes.} 
To evaluate attacks on datasets with continuous attribute values, we compute the \textit{mean squared error} (MSE) between the original nodes and the newly inferred nodes in the candidate set. The mean MSE of all the nodes in the candidate set is then reported. A lower MSE indicates a more successful attack.

%%%%%%%%%%%%%%%%%%%%%%%%%%%%%%%%%%%%%%%%%%%%%%%%%%%%%%%%%%%%%%%%%%%%%%%%%%%%%%%%
\end{document}